\documentclass{article}

\usepackage[preprint]{neurips_2021}

\usepackage[utf8]{inputenc} 
\usepackage[T1]{fontenc}    
\usepackage{hyperref}       
\usepackage{url}            
\usepackage{booktabs}       
\usepackage{amsfonts}       
\usepackage{nicefrac}       
\usepackage{microtype}      
\usepackage{xcolor}         
\usepackage{graphicx}
\usepackage{amsmath,amsthm}

\usepackage[title]{appendix}

\newtheorem{theorem}{Definition}

\usepackage{algorithm}
\usepackage{algpseudocode}
\usepackage{subcaption}
\title{Grid Partitioned Attention: Efficient Transformer Approximation with Inductive Bias for High Resolution Detail Generation}

\usepackage[normalem]{ulem}
\usepackage{cancel}

\newcommand\soutpars[1]{\let\helpcmd\sout\parhelp#1\par\relax\relax}
\long\def\parhelp#1\par#2\relax{%
  \helpcmd{#1}\ifx\relax#2\else\par\parhelp#2\relax\fi%
}
\newcommand{\ttimes}{\hspace{-0.3em}\times\hspace{-0.3em}}

\newcommand{\Fp}{\mathcal{F}} 
\newcommand{\Fc}{\mathcal{F}^{-1}} 

\author{%
  Nikolay Jetchev, Gökhan Yildirim, Christian Bracher, Roland Vollgraf   \\
  Zalando Research\\
  Zalando SE\\
  Berlin, Germany \\
  \texttt{nikolay.jetchev@zalando.de} \\
   \And
   \\
   \And
    \\
    \AND
   \\
}

\begin{document}

\maketitle

\begin{abstract}
Attention is a general reasoning mechanism than can flexibly deal with image information, but its memory requirements had made it so far impractical for high resolution image generation.
We present Grid Partitioned Attention (GPA), a new approximate attention algorithm that leverages a sparse inductive bias for higher computational and memory efficiency in image domains: queries attend only to few keys, spatially close queries attend to close keys due to correlations. 
Our paper introduces the new attention layer, analyzes its complexity and how the trade-off between memory usage and model power can be tuned by the hyper-parameters. 
We will show how such attention enables novel deep learning architectures with copying modules that are especially useful for conditional image generation tasks like pose morphing.
 Our contributions are (i) algorithm and code\footnote{\url{https://github.com/zalandoresearch/gpa}} of the novel GPA layer, (ii) a novel deep  attention-copying architecture, and (iii) new state-of-the art experimental results in human pose morphing generation benchmarks.
\end{abstract}

\section{Introduction}
\label{sec:copyapproach}

The latest generative models like Generative Adversarial  Networks (GANs) have impressive capabilities \cite{karras2020analyzing,brock2019large}:
they generate realistic high-resolution human faces; they handle complex class conditioning (e.g.\ ImageNet \cite{imagenet}) or semantic-map conditioning \cite{wang2018pix2pixHD}.
However, we think that these methods need further improvement in the following aspects, especially for the case of generation conditional on other images:
\begin{itemize}
\item accurately represent the high-frequency details (complicated textures) of diverse distributions 
\item deal with rare and unique visual artifacts (e.g. text logo), which may come only at test time
\end{itemize}

\begin{figure}[tb]
\begin{minipage}[b][][b]{0.55\textwidth}
\includegraphics[width=\textwidth]{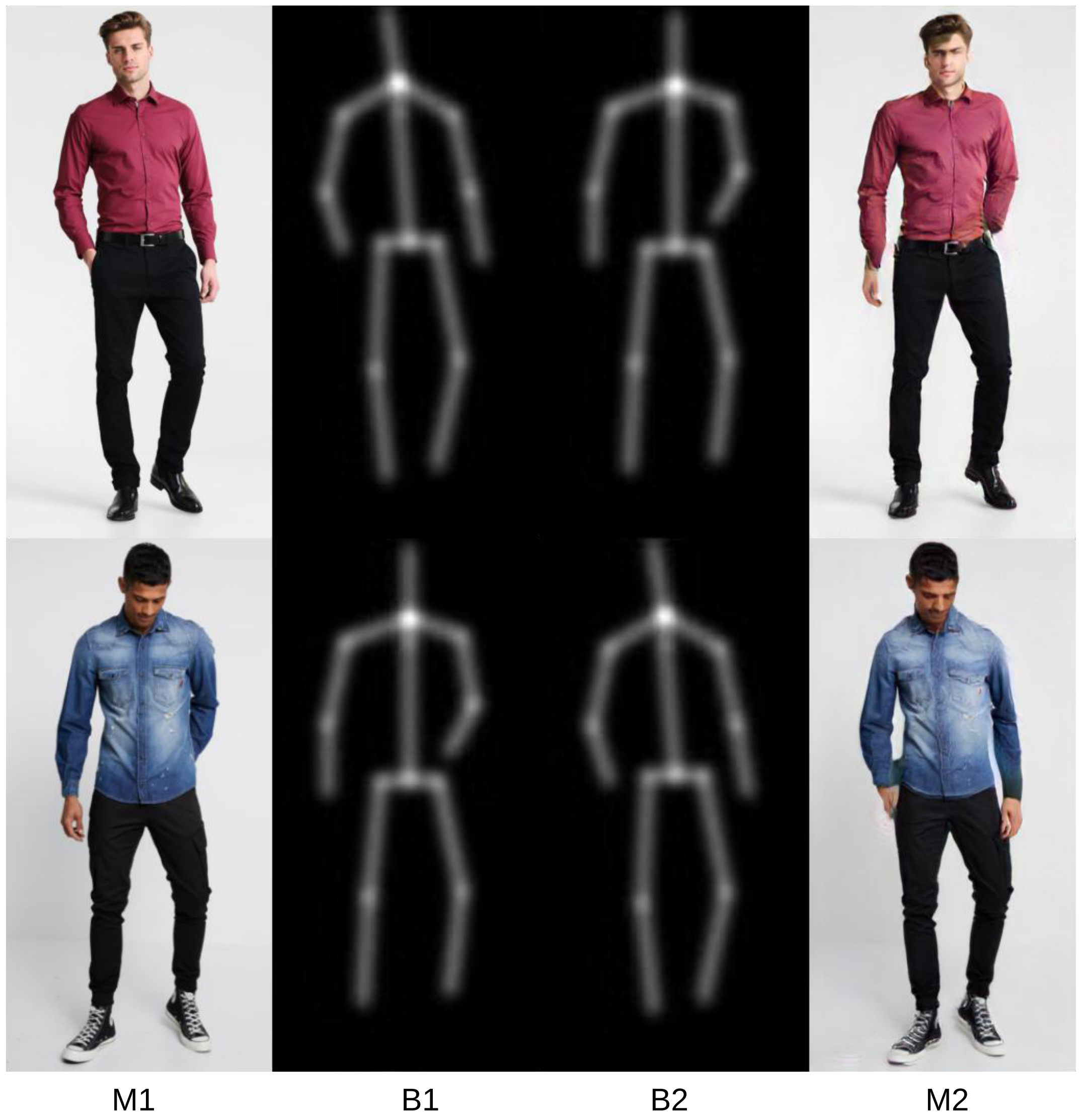}
\end{minipage}
\hfill
\begin{minipage}[b][][b]{0.39\textwidth}
\caption{Example of attention copying for pose transfer: the task is to morph the source appearance $M1$ to target pose $B2$. The novel GPA attention layer can run at high resolutions (here $512\ttimes256$px) and allows generative architectures that efficiently leverage conditioning information and copy important details.}
\label{fig:image1}
\end{minipage}
\end{figure}

The scientific challenge lies in the development of new deep architectures with the properties to (i) learn parametric distributions over data, as traditional generative models,
(ii)  copy flexibly from (strongly warped) conditioning images with minimal loss of the conditioning information, and
(iii) spatially blend the two approaches in a computationally feasible end-to-end differentiable learning framework.
Fig.\ \ref{fig:image1} is an example of the visual quality of our novel copying architecture that has these properties. Its approach is very efficient for image-conditional generation (e.g.\ pose morphing \cite{NIPS2017_posegen}), since we leverage the conditioning information directly through attention, avoid encoding it in a latent embedding space (which requires abundant training
examples) while still benefiting from the generalisation capabilities of a GAN model.



One existing approach to copy-generation relies on appearance flow \cite{flow2016}:
predict flow coordinates and deform the input images before feeding to a convolutional architecture.
An advantage of such warping is that it can copy smooth continuous regions of an image.
However, the extreme locality of flow models makes them difficult to train by standard optimization:
high-frequency gradients destabilize learning \cite{Kanazawa18}, as each copied pixel blends only few neighbouring grid pixels. 

Classical encoder/decoder models (e.g.\ compressing autoencoders) require huge model capacity to reconstruct information. Copying image content (in pixel or feature space) helps (e.g.\ skip connections in UNets \cite{ronneberger2015unet}), but spatial alignment is an issue as convolutions allow only local displacements. Deformable skip connections can help \cite{Siarohin_2018_CVPR}, but training is tricky due to unstable warping flow. 
\cite{Siarohin_2019_NeurIPS} generates by deforming and copying from a source image -- this is a way to recreate image details, without needing model capacity to learn all such data distribution details from training data. 
A similar approach was followed in \cite{jetchevmagic19} -- given a set of input images, the model learns to reuse parts of the input, deform them and blend them in a convincing final image. 
However, this works best only with spatially aligned semantic image regions.

Attention \cite{vaswani2017attention} can be considered as a more general alternative to appearance flow for content copying. It produces an affinity map 
over the whole spatial extent of the key/value input that weights the contribution of each input pixel to the total output. 
Attention mechanisms exhibit highly non-local behaviour and copy information from multiple locations at once. This can lead to better gradient training than flow warping. 
In contrast, we can see appearance flow as attention in the limit where the whole affinity mass for a given query is concentrated at a single key/value. 
In addition, attention can flexibly integrate geometric inductive bias for the affinity map, e.g.\ enforce smoothness or leverage coordinates in keys and queries as inductive bias. 

The downside of using attention is the memory cost, 
e.g.\ for $128\ttimes128$ pixel images, the query/key tensors have $2^{14}$ elements, and attention needs an affinity matrix with $2^{28}$ elements, too big for current GPUs.
An approximation is required that balances the expressive power of the model against its memory (and computation) costs.
In this paper we propose {\em Grid Partitioned Attention (GPA)}, a novel attention algorithm with such properties.
Sec.\ \ref{sec:GPA_whole} will describe in detail the GPA algorithm.
Sec.\ \ref{sec:Exp} will demonstrate experimentally how GPA enables novel deep architectures, which generate high resolution images with improved detail reconstruction. 
We focus on {\em pose morphing}, re-rendering the image of a person in a different pose, as a good example of stringently conditioned image generation.

\section{The GPA attention layer}
\label{sec:GPA_whole}
\subsection{Notation}
\label{sec:notation}

\begin{figure}[tb]
\centering
\includegraphics[width=0.85\textwidth]{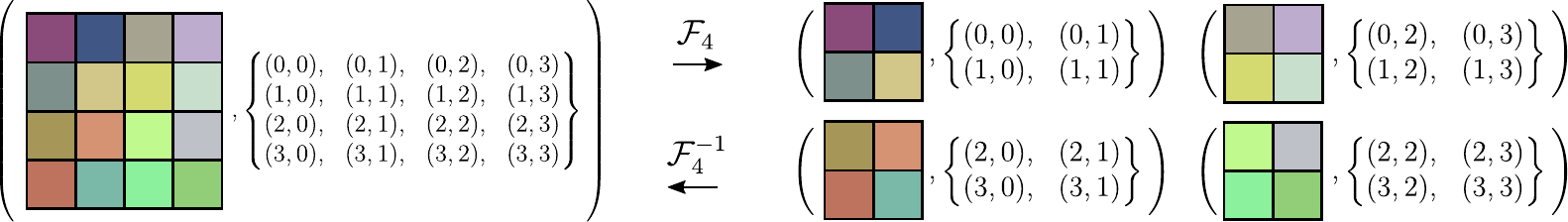}
\caption{Illustration of spatial partitioning  (Definition \ref{def:partition}) and composition (Definition \ref{def:composition}) operations.}
\label{fig:upd}
\end{figure}

Images are represented usually as tensors $X \in \mathbb{R}^{c\times h\times w}$. The 3 dimensions are channel,  and two spatial dimensions (height and width).
We address individual pixels of an image with their spatial coordinates defined with index sets $I\subseteq I_{hw}=[0,h-1]\times[0,w-1]\subset\mathbb N^2$, where $I_{hw}$ is the full spatial index structure of tensor $X$.
However, we can also have index arrays $I' \subset I_{hw}$ with less elements, indexing an image $X'$ cropped out of tensor $X$. Spatial index structures may exhibit a spatial structure on their own, in which case we would define $I_{hw}\in\left([0,h-1]\times[0,w-1]\right)^{h\times w}$ rather as a tensor than a set. Which interpretation is used shall be explicitly mentioned or be clear from the context.
Image tensors allow for simple indexing where we implicitly slice over the channel dimension. Thus, $X_{i,j}=X_{(i,j)}\in\mathbb R^c$ and $X_I=\left\{X_{(i,j)}\right\}_{(i,j)\in I}$. If $I$ has a spatial structure, then $X$ is a tensor with the same spatial structure. 

We start with a formal definition of several operations required to implement GPA.
\begin{theorem}[Downsampling image tensors]
Let $X\in \mathbb{R}^{c \times h\times w}$ be an image tensor and (for simplicity and w.l.o.g.) $h,w$ be multiples of $d\in\mathbb N^+$. We define downsampling with factor $d$ as 
$D_d(X):=X' \in \mathbb{R}^{c \times \frac{h}{d}\times \frac{w}{d}}$ where $X'_{i,j} = \frac{1}{d^2}\sum_{\delta_h=0}^{d-1}\sum_{\delta_w=0}^{d-1} X_{d \cdot i+\delta_h,d \cdot j+\delta_w}$.
\end{theorem}

\begin{theorem}[Upsampling index sets]
Let $I'\subseteq I_{h'w'}$ be a spatial index structure. We define upsampling of $I'$ with factor $d$ as $U_d(I'):=I=\left\{(i'\cdot d+\delta_h,j'\cdot d+\delta_w)\right\}_{(i',j')\in I',\delta_h\in[0,d-1],\delta_w\in[0,d-1]}$. 
\end{theorem}

Per definition of $\mathcal{D}_d$ and $\mathcal{U}_d$, pixels of downsampled images are averages over the upsampled indices $X'_{i,j} = \frac{1}{d^2}\sum_{(k,l)\in \mathcal{U}_{d}\left(\{(i,j)\}\right)}X_{k,l}$.

\begin{theorem}[Spatial partitioning]
\label{def:partition}
Let $X$ be a tensor with spatial index structure $I$. An operation  
$\Fp_m(X):=\left\{\left(X^{(l)},J^{(l)}\right)\right\}_{l=1\ldots m}$ where $X^{(l)}=X_{J^{(l)}}$ and $\bigcup_l J^{(l)}=I$ and $J^{(l_1)}\cap J^{(l_2)}=\emptyset$ for $l_1\neq l_2$ is called spatial partitioning of $X$ into $m$ partitions (cells).
\end{theorem}
Usually, spatial partitioning is implemented such that the $J^{(l)}$ have a spatial structure and, hence,  $X^{(l)}$ are tensors. 
\begin{theorem}[Spatial composition]
\label{def:composition}
Let $\mathcal Y = \left\{\left(X^{(l)},J^{(l)}\right)\right\}_{l=1\ldots m}$ be a spatial partitioning. If there is exactly one $X$ such that $\Fp_m(X)=\mathcal Y$, then $\Fc_m(\mathcal Y):=X$ is the unique spatial combination of $\mathcal Y$.
\end{theorem}

\begin{theorem}[Consistent spatial partitioning]
Let $X$ be an image tensor, and $X' = \mathcal D_d(X)$.
Let $\Fp_m$ be a spatial partitioning and $\left\{\left(X^{(l)},J^{(l)}\right)\right\}_{l=1\ldots m}=\Fp_m(X)$ and $\left\{\left(X'^{(l)},J'^{(l)}\right)\right\}_{l=1\ldots m}=\Fp_m(X')$.
$\Fp_m$ is said to be consistent, iff\quad $\mathcal{U}_d(J'^{(l)}) = J^{(l)}$  for all $l=1\ldots m$.
\end{theorem}
For example, clipping an (image) tensor into equally sized pieces (cf.\ Fig.\ \ref{fig:upd}), which are arranged in a specific order (e.g.\ scan line order) is a consistent spatial partitioning. 
\subsection{GPA algorithm}
\label{sec:gpa}

Suppose we have flattened query, key and value tensors $Q,K \in \mathbb{R}^{c_k \times n}$, $V \in \mathbb{R}^{c_v \times n}$, where $n=hw$, and want to calculate attention 
\begin{align}
    \label{eq:attention}
    X= \mathcal{A}(Q,K,V)=V\cdot\mathcal{S}(Q,K)=V\cdot\mathrm{softmax}(K^\top Q) .
\end{align}

This operation is expensive in memory due to the $n \times n$ matrix $A=\mathcal{S}(Q,K)$.
We will make three useful assumptions for the image domain as inductive bias for the approximation of GPA.
\textbf{A1} local spatial correlation: natural images have a structure where close positions are correlated (a counterexample is a white noise image, where there is no correlation) \cite{Cecchi}\cite{Maheswaranathan340943}.
\textbf{A2} scale correlation: assume that the attention structure at lower scale roughly corresponds to the attention structure at high scale \cite{Ruderman1994}. 
\textbf{A3} attention sparsity: for any query, only a fraction of the keys have non-zero affinity values, as in dense correspondence methods \cite{rocco2018neighbourhood} and recent sequence transformer models \cite{bigbird}.

\begin{figure}[tb]
\centering
\includegraphics[width=0.8\textwidth]{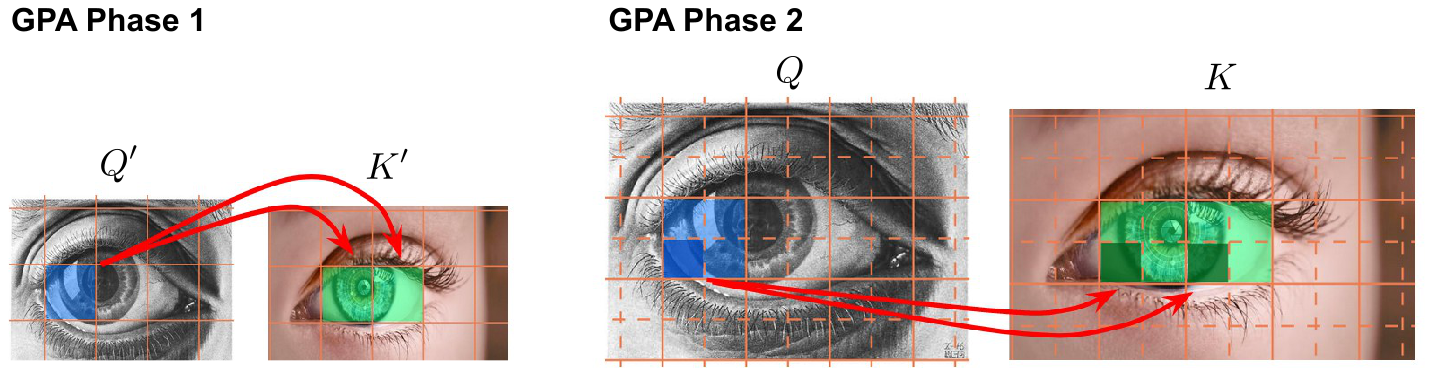}
\caption{Illustration of GPA: how to calculate very sparse attention from query tensor $Q$ to keys $K$. Phase 1: downsample into $Q',K'$ and find keys that have the largest affinity for a query cell -- the relevant key set. We highlight in blue a query position, and in green its $\kappa=2$ top keys.
Phase 2: given the keys and queries from Phase 1, we can up-sample their indices and get an approximation for the relevant key set in the original size tensors $Q,K$.
}
\label{fig:p12}
\end{figure}

Let $J$ be the indices of the queries (column indices of $Q$) and $I$ the indices of the keys (column indices of $K$).
For a single query $Q_j$ we have that $A_j = \mathcal{S}(Q_j,K)$ which takes $O(n)$ memory.
If we can identify a subset of key indices $I(j)\subseteq I$ that are relevant for the query $Q_j$, we can calculate just the attention with respect to those keys and realize great computational savings if $|I(j)|\ll|I|$ (see assumption \textbf{A3}). 
How can we find $I(j)$ in a computationally efficient way? A simple greedy relevance algorithm is to take the $\kappa$ key rows of $A$ with highest average activation over all query columns. 
Formally, $\mathcal{R}_{\kappa}(A,J) := I\kappa(J)$ where $|I_\kappa(J)|=\kappa$ and $i\neq i'$ for all $i,i'\in I\kappa(J)$ and $\sum_{j\in J} A_{i,j} \ge \sum_{j\in J} A_{i^*,j}$ for all $i\in I\kappa(J), i^*\in I\setminus I\kappa(J)$.

Trivially, we could calculate $R_\kappa(A_j, \{j\})$ where $A_j = \mathcal{S}(Q_j,K)$ and keep just $\kappa$ elements of $A_j$ for subsequent attention calculation. However, this would be quite expensive and saves us no memory at all if we need to calculate the full attention in order to sort all elements by magnitude.
Motivated by assumption \textbf{A2}, we downsample the keys and queries into $Q'=\mathcal{D}_d(Q)$ and $K'=\mathcal{D}_d(K)$ by factor $d$, and then, at the lower scale, calculate the full affinity matrix $A'= \mathcal{S}(Q',K')$.
The relevant key sets at the lower level can help us approximate the relevant sets at the higher level, see Fig.\  \ref{fig:p12} for illustration.
However, simply computing $\mathcal{R}_{\kappa}(A',J_{h'w'})$ would have no other effect than setting the key-value dictionary size to $\kappa d$ elements globally for all queries. This is not exactly what we want. Instead, following assumption \textbf{A1}, we perform this step individually for different local neighborhoods of the image, the spatial partition cells (Definition \ref{def:partition}).




We describe the GPA algorithm as Algorithm \ref{alg:algorithm}.
$\Fp_m$ is usually the square cell partitioning function, which is fast and available in many deep learning frameworks. 
In Phase 1 we find the relevant key sets at a low resolution $h' \times w'$ with $h'=\frac{h}{d},w'=\frac{w}{d}$, for which full attention is still feasible in memory. This is not done for single query elements, but more generally for consistent spatial partitions of query locations. By assumption \textbf{A1} keys and queries are spatially correlated: if $Q_{j_1}$ attends to keys at index set $I_1$, and $Q_{j_2}$ attends to the keys at index set $I_2$, then for $j_1$ and $j_2$ close to each other, the relevant key index sets $I_1,I_2$ will be similar. Here, the partition parameter $m$ is a hyper-parameter that reflects the level of smoothness according to  \textbf{A1}. 
In Phase 2 we leverage the relevant key set at the high resolution.
On line \ref{alg:upsample}, we use the index upsampling operator $\mathcal{U}_d$ to sample around the relevant key sets from the low spatial level -- we densely fill the finer grid around the locations we choose. Other sampling strategies are possible, but this one is fast to implement and easy to parallelize.

\newcommand{\cuscomment}[1]{\State{\color{blue}{{\em  \#\# #1  \#\#}}}}
\begin{algorithm}[tb]
\caption{Grid Partitioned Attention (GPA)} 
\label{alg:algorithm}
\begin{algorithmic}[1]
\State Input: $Q,K\in\mathbb R^{c_k\times h\times w},V\in\mathbb R^{c_v\times h\times w} $
\Comment{Query, Keys and Values}
\State\phantom{Input:}$\Fp_m$
\Comment{consistent spatial partitioning} 
\State Parameters:  $m$ count of partition cells, $d$ downsampling factor, $\kappa$ size of relevant key set
\cuscomment{Phase 1: finding relevant keys at the low resolution}
\State $Q' = \mathcal{D}_d (Q),K' = \mathcal{D}_d (K)$
\Comment{down-sample by factor $d$, so that $Q',K' \in \mathbb{R}^{c_k \times h'\times w'}$}
\State $A' = \mathrm{softmax}\left((K')^\top Q'\right)$
\Comment{reshape $Q',K'$ to size $c_k\times h'w'$; matrix $A'\in \mathbb{R}^{h'w' \times h'w'}$,  }
\State reshape $A'$ into size $h' \times w'  \times h'  w' $ \Comment{the queries dimension of $A'$ is reshaped spatially}
\State  $\left\{\left(A'^{(l)}, J'^{(l)}\right)\right\}_{l=1}^m = \Fp_m(A')$ 
\Comment{partition the query spatial dimensions of $A'$ into $m$ cells}
\For{$l=1$ \dots $m$}
            \State $I'^{(l)} = \mathcal{R}_{\kappa} (A'^{(l)},J'^{(l)})$
            \Comment{find the $\kappa$ most relevant keys for each cell of grouped queries}
\EndFor
\cuscomment{Phase 2: using the relevant keys for approximate attention at the original resolution}
\State $\left\{\left(Q^{(l)},J^{(l)}\right)\right\}_{l=1}^m = \Fp_m(Q) $ \label{alg:partition}
\Comment{partitions are related across scales $J^{(l)} = \mathcal{U}_d(J'^{(l)})$}
\For{$l=1$ \dots $m$}
\State  $I^{(l)} = \mathcal{U}_d(I'^{(l)})$\label{alg:upsample}
\Comment{indices corresponding to the low-res key set $I^l$}
\State gather subtensors $K_{I^{(l)}},V_{I^{(l)}}$
\Comment{tensors of size $\mathbb{R}^{c_k \times d\kappa},\mathbb{R}^{c_v \times d\kappa}$}
\State  $X^{(l)} = \mathcal{A}\left(Q^{(l)},K_{I^{(l)}},V_{I^{(l)}}\right)$
\Comment{$X^{(l)}$  is a subset of the GPA output 
}
\EndFor
\State $X=\Fc_m\left(\{X^{(l)}, J^{(l)}   \}_{l=1}^m\right)$
\Comment{combine full size attention output from the query cells}
\end{algorithmic}
\end{algorithm}

\label{sec:complexity}
Complexity is $\mathcal{O}(\frac{n^2}{d^2})$ for Phase 1, due to a call to full attention after downsampling.
For Phase 2 the memory cost is $\mathcal{O}(n d^2 \kappa )$, where for each partition we call $m$ times attention between $\frac{n}{m}$ queries and $d^2 \kappa$ keys in the relevant key set.%
\footnote{All $m$ attention calls will be  executed efficiently in parallel on the GPU.}
If we set $\kappa=\frac{n}{d^2}$, then the approach becomes equivalent to full attention, since at the original scale all $n$ keys are considered for all queries.
If we set $m=1$, then all queries are in the same partition, and $\kappa$ tunes how many keys to attend to globally.

\subsection{Related works: approximate attention}
Full attention is used successfully in transformer architectures for vision, language, and reinforcement learning, but its memory footprint often limits the tensor sizes it can process.
A lot of work deals with approximations, which can be much lighter but introduce an error, see \cite{anonymous2021long,tay2020efficient} for an overview. 
GPA will be placed at the intersection of fixed and learnable patterns for attention.
It uses a pattern for the queries that is regular and fixed w.r.t. the image grid,  while extracting the most relevant keys (via sorting) is an adaptive yet non-differentiable approach. 
GPA differs from random sparse attention patterns such as \cite{bigbird}, because we use spatial grid partitioning as inductive bias and leverage sparsity in many query neighborhood cells.

Ref.~\cite{kitaev2020reformer} uses Locally Sensitive Hashing (LSH) to sort the keys/queries, assuming that the key partitions are of similar size and non-overlapping.
In contrast, GPA deals with relevant keys without any such restrictive  assumptions: it finds the relevant keys for a query partition independently of the keys for all other partitions.
In addition, for the applications we have in mind (e.g. conditional GAN models) it holds that queries, on one side, and keys and values, on the other side, come from different information pathways -- GPA handles that case naturally. In contrast,  ref.~\cite{kitaev2020reformer} is specifically designed for self-attention problems where keys and queries share the same information. 

Ref.~\cite{vyas2020fast} clusters the queries and computes attention for the centroids, while keeping all keys.  This is a costly strategy for high-resolution images. Clustering also adds an approximation error since for all cluster members (sharing the same centroid) attention is exactly the same.  
GPA partitions the queries too, but we handle sparsity differently, and each query in a partition can get a different attention.




\section{Experiments}
\label{sec:Exp}

We ran experiments on the popular DeepFashion (DF) dataset \cite{d_fash}, using the standard training and test splits \cite{NIPS2017_posegen}, and 18 pose keypoints as in \cite{zhu2019progressive}. Image size is $256\ttimes176$, which we pad to a square when training. DF allows a clean quantitative comparison with many other papers using this dataset setup.
For qualitative comparison, in Sec.\ \ref{sec:highres} we run additional  experiments on two larger, higher resolution datasets. Their rich image distributions give more insight into the benefits of detail copying. 

\subsection{The copying attention: architecture and loss}
\label{sec:archiloss}

\begin{figure}[t]
\centering
\includegraphics[width=12cm]{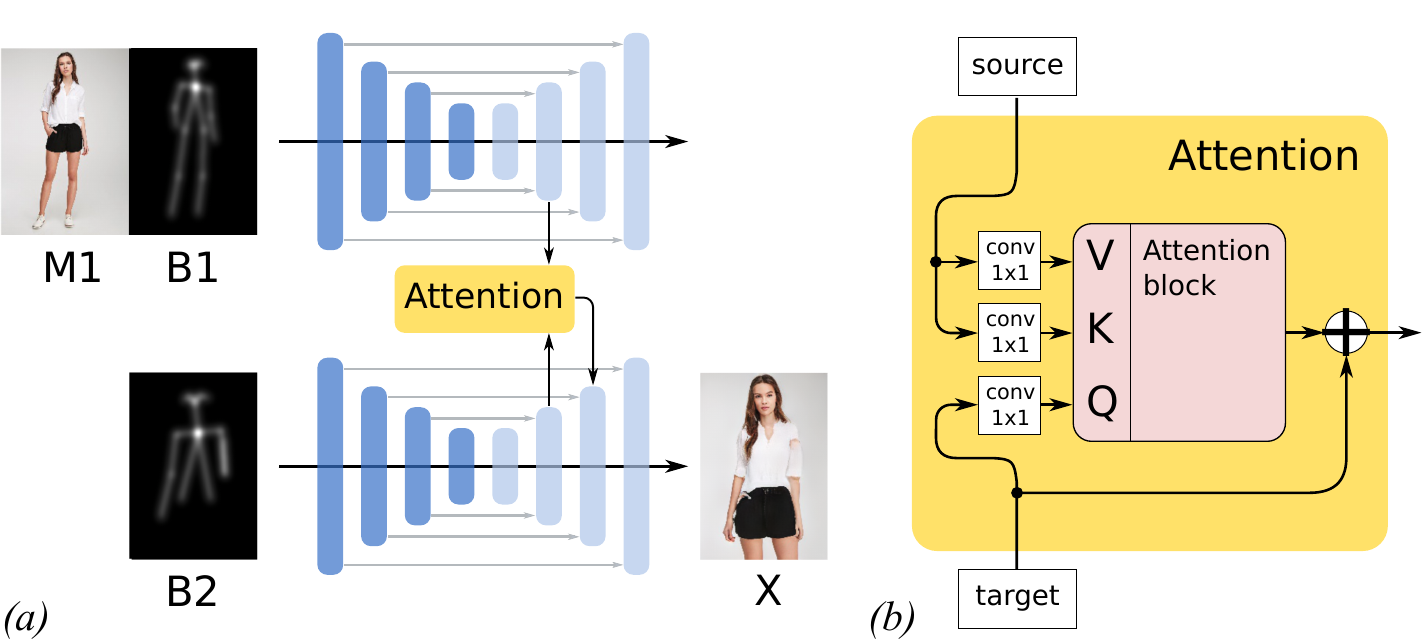}
\caption{ \textit{(a)} Scheme of pose-morphing generator architecture $G$ with two UNets: one for the concatenated source appearance (M1) and pose (B1) and one for the target pose (B2). The later outputs the generated target appearance ($X$). \textit{(b)} Attention blocks copy information from the source decoder layers (keys, values) to the appropriate spatial positions in the target decoders (query).}
\label{fig:posearchi}
\end{figure}

A generator for the pose-morphing task can be considered as a function $G$ which takes as input a source image $M_1$, its pose $B_1$, and a target pose $B_2$, and produces an image $X$. The training loss (defined below) is designed to make $X$ the likeness of $M_1$ but change pose to $B_2$  -- hence "pose-morphing".
Fig.\  \ref{fig:posearchi}a) shows our architecture with attention copying. The key/value tensors (for the attention) come from the decoder blocks of the source appearance and pose UNet \cite{ronneberger2015unet} (featuring encoder and decoder with skip connections), and the query tensors come from the decoder of the target branch UNet which generates the output image.  All convolutional blocks are designed as in \cite{karras2020analyzing}. 
Typically we used 6 blocks in each encoder and decoder (see Appendix for details). The poses are represented as 18 keypoint channels, each being a heatmap around a pose keypoint. For ease of interpretation, our figures show skeletons connecting these keypoints.


Fig.\  \ref{fig:posearchi}b) shows the details of the main novel component of the architecture: an attention layer that efficiently leverages and ``copies'' image content. It has 3 trainable $1\ttimes1$ convolutions: for key and value (source what to copy), and the query (target where to copy). Given these, the attention module (GPA or other attention type) will propagate information from the value tensor, according to key-query affinity.
We can either add as residual to the generated feature map, or concatenate the attention output as additional channels, which we used in our experiments.
This design allows the network to learn complex copying behaviour. For DF data, we define the \textbf{pivot} configuration: use full attention layers for tensors with size $64\ttimes64$ or less (which fit in memory for that size), and for larger sizes ($128\ttimes128$ and $256\ttimes256$) use approximate attention layers.
All GPA layers downsample to size $64\ttimes64$ for Phase 1. At that size, queries are partitioned into $m=32 \ttimes 32$ cells, each of spatial size $2 \ttimes 2$. We used $\kappa=12$ keys for each cell. Phase 2 works on the original large sizes, respectively.   
We also tried multihead versions of GPA (and other attention blocks), but found no performance gain despite more memory usage; thus we used single head for the final experiments.

We train the pose-morphing network with a loss combining 3 terms:
\begin{align}
\mathcal{L} =\mathbb{E}_{(M_1,B_1,M_2,B_2)} \left[ \mathcal{L}_{per}(M_1,B_1,M_2,B_2)+\mathcal{L}_{self}(M_1,B_1) + \mathcal{L}_{GAN}(M_1,B_1,B_2) \right]
\end{align}
For datasets (e.g.\ DF) with pairs of source $M_1,B_1$ and target $M_2,B_2$, we can directly use a \textbf{perceptual loss} \cite{Zhang_2018_CVPR} using paired data $\mathcal{L}_{per}(M_1,B_1,M_2,B_2)=\|G(M_1,B_1,B_2)-M_2  \|_{vgg}$. The error metric uses features from the pretrained VGG network \cite{vgg}.
For any dataset we can use also a \textbf{self-supervised loss} \cite{jing2019selfsupervised} by creating pairs of frames $M,B,\tau(M),\tau(B)$, where $\tau$ is a spatial transformation applied on both image and pose. We define $\mathcal{L}_{self}(M,B) = \|G(M,B,\tau(B))-\tau(M)\|_{vgg}$, which forces the generator to reproduce the target image. Copying details is an essential mechanism for that.
In general, $\tau$ can be designed in many ways (see  \cite{jing2019selfsupervised}), but 
we opted for a simple affine deformation. 
To randomly sample affine transformation matrices we added noise $\epsilon \sim \mathcal{U}(-0.15,0.15)$ to the identity transform. 
While most pose changes in data are more complex than affine transforms, the  self-supervised loss was helpful and stabilized GAN training behaviour.  
We use also a standard \textbf{GAN loss}, with a discriminator that sees two types of pairs: true data $(M_1,B_1)$ and generator output $(G(M_1,B_1,B_2),B_2)$, where $B_2$ is the pose the image should have.
The discriminator is sensitive both to the quality of the image, and its match with the chosen pose. The discriminator and loss had the StyleGAN design \cite{karras2020analyzing}. 
We optimized the loss with ADAM \cite{kingma2017adam} over minibatches of 3 images for $256\ttimes256$ pixel images, on  Nvidia V100 GPUs in our internal cluster, in PyTorch \cite{paszke2019pytorch}. All our GAN models were trained for 7 days on one GPU core; the supervised distortion task ran for 5 days. 

\subsection{DF benchmark: affine transformation copying and in-person pose morphing}
\label{sec:ssbench}
We want to investigate which attention block works best in the architecture from Fig.\ \ref{fig:posearchi}) for a DF image distortion task. 
For that we used only loss term $\mathcal{L}_{self}$ and fixed training time to 5 days, a fair method comparison.
The experimental loss is measured on a test set of 500 images and random affine transforms.
All configurations used full attention for the 4 decoder blocks until size 64x64.
We tested the \textbf{pivot} configuration of GPA1024\_12\_64 (as in Sec.\ \ref{sec:archiloss}), and several ablations around it. Each is named $\textrm{GPA}m\_\kappa\_s$, indicating  $m,\kappa$ and the largest resolution $s \ttimes s$ where we used full attention (approximate attentions at size above $s \ttimes s$).
We tested two popular recent low rank attention approximation methods, the Nystromformer NYF \cite{xiong2021nystromformer},
and the Performer PER \cite{choromanski2020rethinking}, set up with 512, 708 or 768 as low rank parameter.
NO indicates no attention layer above size 64x64. 

\begin{figure}[t]
        \centering
        \begin{subfigure}[b]{0.32\textwidth}
           \includegraphics[width=\textwidth]{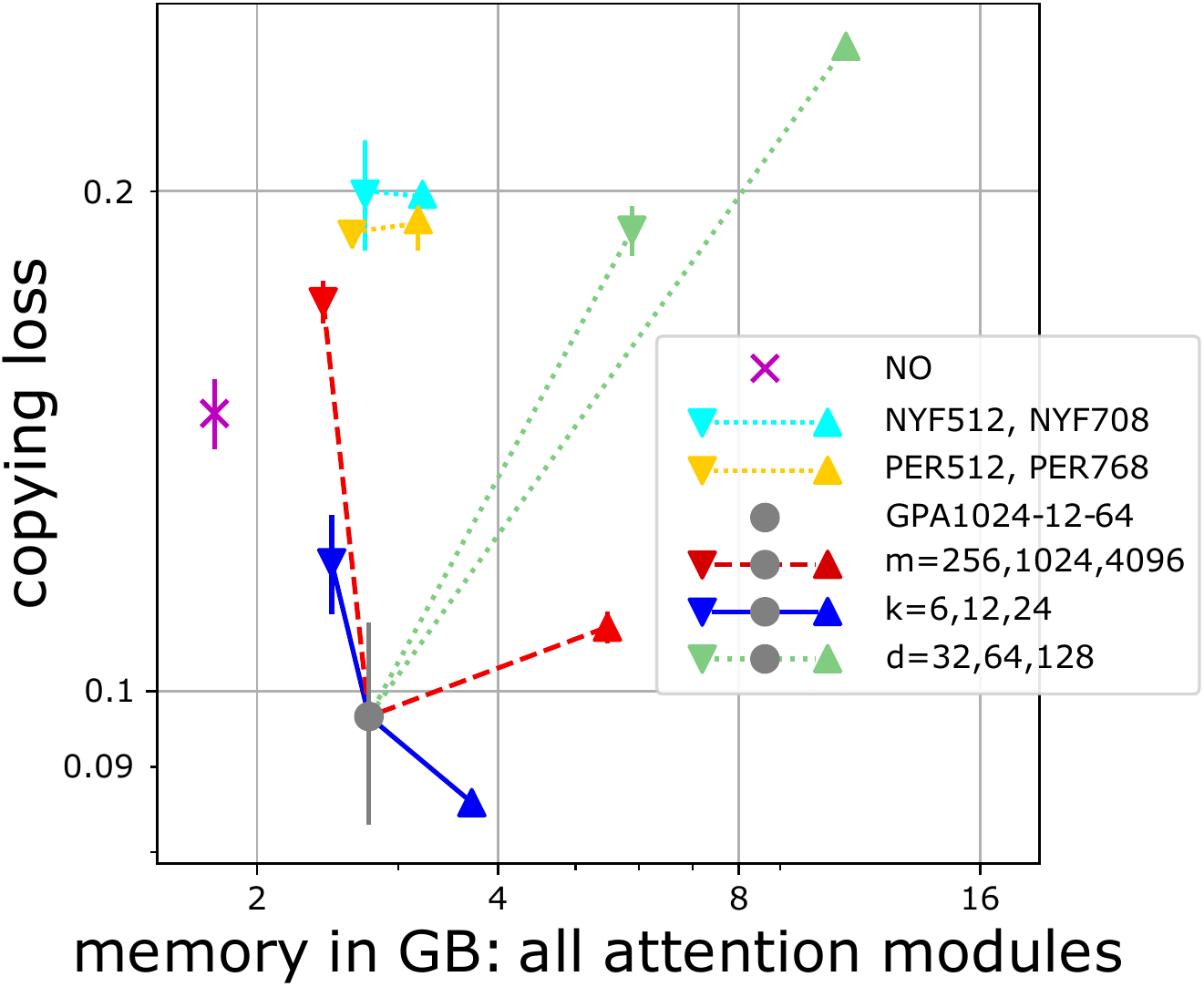}
        \caption{}
        \end{subfigure}
        \begin{subfigure}[b]{0.65\textwidth}
        \includegraphics[width=\textwidth]{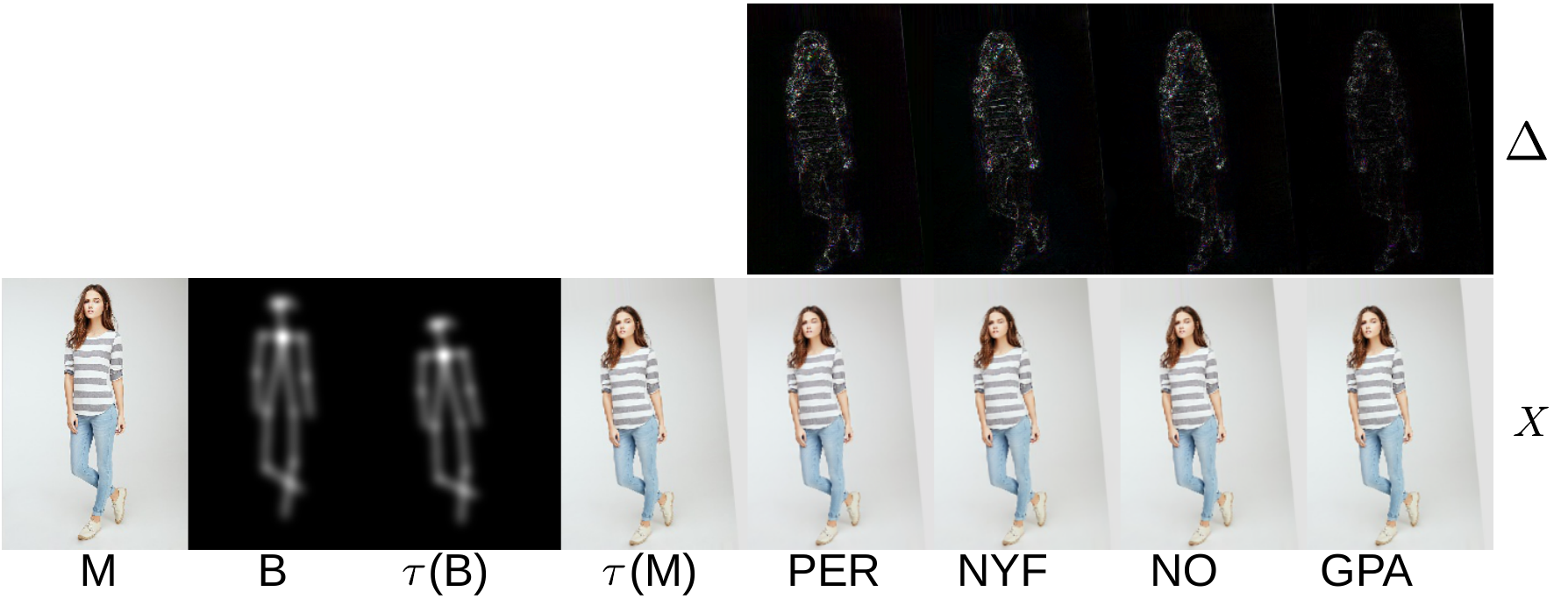}
        \caption{}
        \end{subfigure}
        \caption{Image distortion task: conditional on $M$, $B$, and $\tau(B)$ the network needs to generate $\tau(M)$. We test various attention blocks. \textit{(a)} Attention memory cost vs loss (error bars for 3 runs of 5 days). \textit{(b)} Example generated images $X$ -- complex textures (e.g. stripes) are transferred best by GPA. The top row shows $\Delta = \|\tau(M) -X\|_1$, absolute error of the images $X$ as RGB pixel intensity.}
        \label{fig:ss}
\end{figure}

The results are summarized in Fig.\ \ref{fig:ss}.
The statistics (with error bars over 3 randomly initialized runs) confirm that GPA (around the pivot configuration) had clearly the best copying performance. 
We measure the memory\footnote{In PyTorch: do forward/backward pass, call \texttt{torch.cuda.max\_memory\_allocated()}.} just for the 6 attention blocks, calculated in gigabytes (GB) for input tensors with batch size 1. 
For the pivot configuration, GPA uses less memory than PER and NYF.
Interestingly, NYF and PER do not benefit from extra capacity.
NO is better than them, but still worse than GPA.
Note that the ablations changing the scale for GPA Phase 1 to $32\ttimes32$ and $128\ttimes128$ are both more memory costly (and slower) than the $64\ttimes64$ pivot. This is consistent with the GPA complexity formula from Sec. \ref{sec:gpa}. For GPA1024\_12\_128 we used batchsize 2 (instead of 3) due to memory limits. 

\begin{table}[b]
  \caption{Performance of GPA+attention copying against other baseline models.}
  \label{table:DF}
  \centering
  \begin{tabular}{lllll}
        \hline
   Method    & FID     & SSIM & R2G & G2R \\
   \hline
   Attention Copying + GPA & \textbf{8.86}   & \textbf{0.767} &0.12 & \textbf{0.27}   \\
   Attention Copying + NO & 13.05   & 0.752 &\textbf{0.08} & 0.23   \\
    BFT & 14.5  & 0.766 &0.16 &0.21   \\
    DSC & 19.9  & 0.761 &0.15 &0.15   \\
    XING & 42.8  & 0.756 & 0.12 &0.03   \\
    \hline
  \end{tabular}
\end{table}



We performed the classical DF benchmark, with standard train/test splits \cite{NIPS2017_posegen}.
DF has thousands of appearances, several images per appearance. For the in-person pose morphing, images and poses $M_1,B_1,M_2,B_2$ from the same appearance are used.
We compared results using statistical scores (FID \cite{fid} and SSIM \cite{ssim}).
We also did a human perceptual evaluation using 55 true and 55 generated images. The task was to see a random image for 4 seconds (as in \cite{HYPE}), and answer whether it looks real or generated.
We used 3 clicks per image, 330 clicks total, for calculating R2G (percentage of real images considered generated) and G2R (generated considered real). 

\begin{figure}[tb]
\centering
\includegraphics[width=0.85\textwidth]{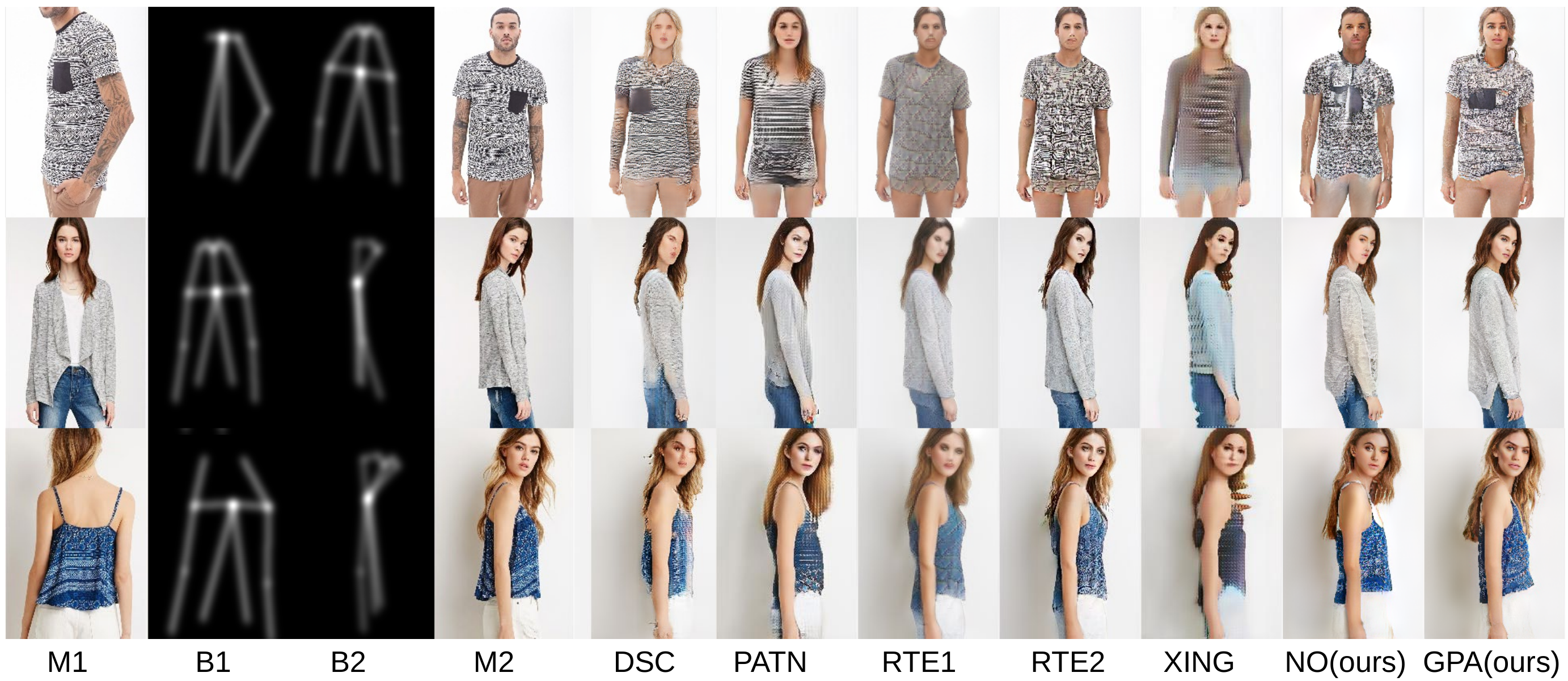}
\caption{Visual comparison of several architectures for the task of generating the appearance of $M_1$ into pose $B_2$, on the DF dataset. GPA (right) generates fine details, e.g. the pocket in the top row.}
\label{fig:dfresult}
\end{figure}

We compare GPA attention copying with NO copying configuration, and BFT \cite{AlBahar_2019_ICCV}, DSC \cite{Siarohin_2018_CVPR}, XING \cite{tang2020xinggan}. Table \ref{table:DF} summarizes our results -- GPA with attention copying is a new state-of-the-art for the DF posemorphing task (FID and crowdsourcing scores are more relevant quality measures than SSIM, as noted by \cite{zhu2019progressive}).
In Fig.\  \ref{fig:dfresult}  we show a visual comparison for several example generated images. The other methods in that figure are PATN \cite{zhu2019progressive} and RTE \cite{Yang2020RegionadaptiveTE}.
 XING was tested using a pretrained model, other reference images were taken from \cite{Yang2020RegionadaptiveTE}.
 We do not report results for \cite{Siarohin_2019_NeurIPS} since we could not train it well on DF data - we suppose that  DF is very different than the training data setup of \cite{Siarohin_2019_NeurIPS}. 


\subsection{High resolution results}
\label{sec:highres}

\begin{figure}[b]
\centering
\begin{subfigure}[b]{0.35\textwidth}
    \includegraphics[width=\textwidth]{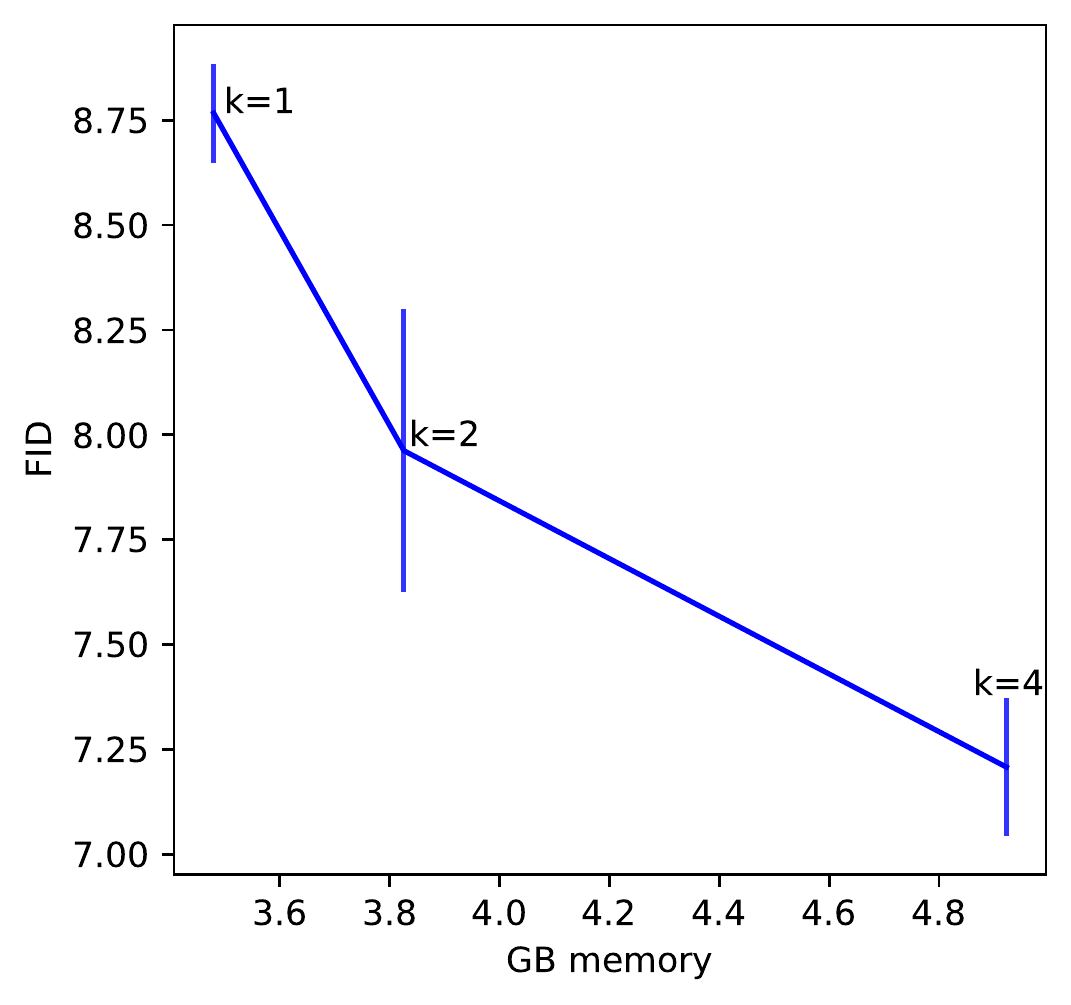}
    \caption{}
\end{subfigure}
\begin{subfigure}[b]{0.64\textwidth}
    \includegraphics[width=\textwidth]{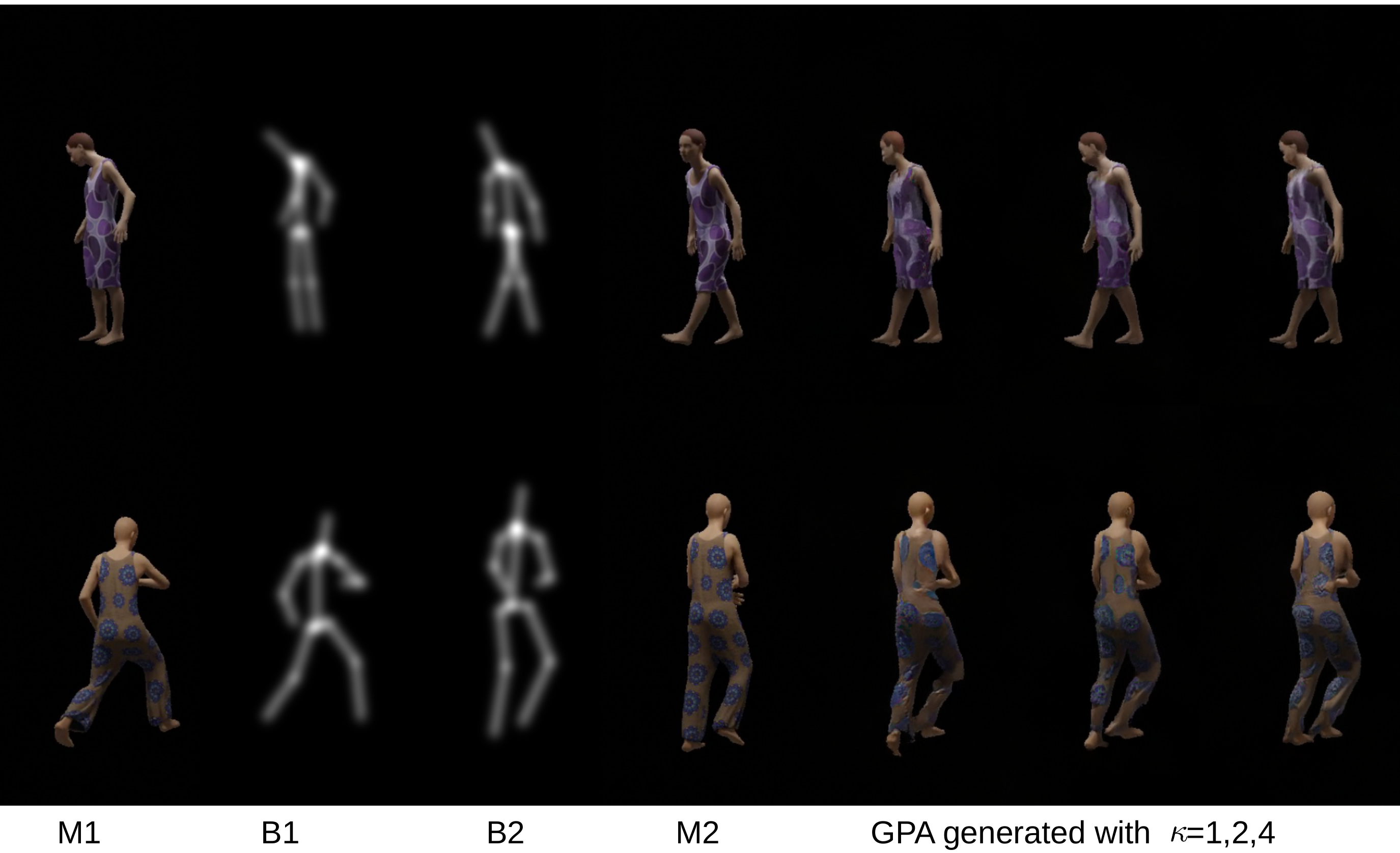}
    \caption{}
\end{subfigure}
\caption{Pose morphing ChaL $512\ttimes256$ pixels. \textit{(a)} Increasing the $\kappa$ parameter of GPA leads to more expensive and powerful attention copying models (error bars over 3 random runs). \textit{(b)} The generated images sorted from left to right with increasing $\kappa$, which correlates with higher visual quality.}
\label{fig:kapabl2}
\end{figure}

We also worked with another high resolution publicly available dataset ChaL\cite{ChaLearn,ChaLearn0}. It has 40.000 videos, 120 frames each -- data for pose morphing between frames of the same appearance.
We extracted keypoints by \cite{HRnet19}, cropped each video to a region of size $512\ttimes256$ pixels around the human figure, and trained our model for 7 days on a V100 GPU with minibatch size 2.
Until size $64\ttimes32$ we used full attention, and for tensors of size $128\ttimes64$, $256\ttimes128$, $512\ttimes256$ we used GPA.
For Phase 1 the query and keys were downsampled to $64\ttimes32$ pixels, and the query was partitioned in $m=64\ttimes32=2048$ pixel-sized cells.
Fig.\  \ref{fig:kapabl2} shows results when training with different number of relevant keys for each downsampled query block. We ablated $\kappa=1,2,4$, and we see how this affects visual fidelity.
As discussed already in Sec.\ \ref{sec:gpa}, the larger $\kappa$ gets, the more capacity the GPA gets -- this correlates also well with FID score, and memory usage. 
The inductive bias of GPA allows it to have good image generation performance, while still being much cheaper (memory and computationally) than full attention, which at this resolution cannot fit in GPU memory.

\begin{figure}[t]
        \centering
        \includegraphics[height=3.35cm]{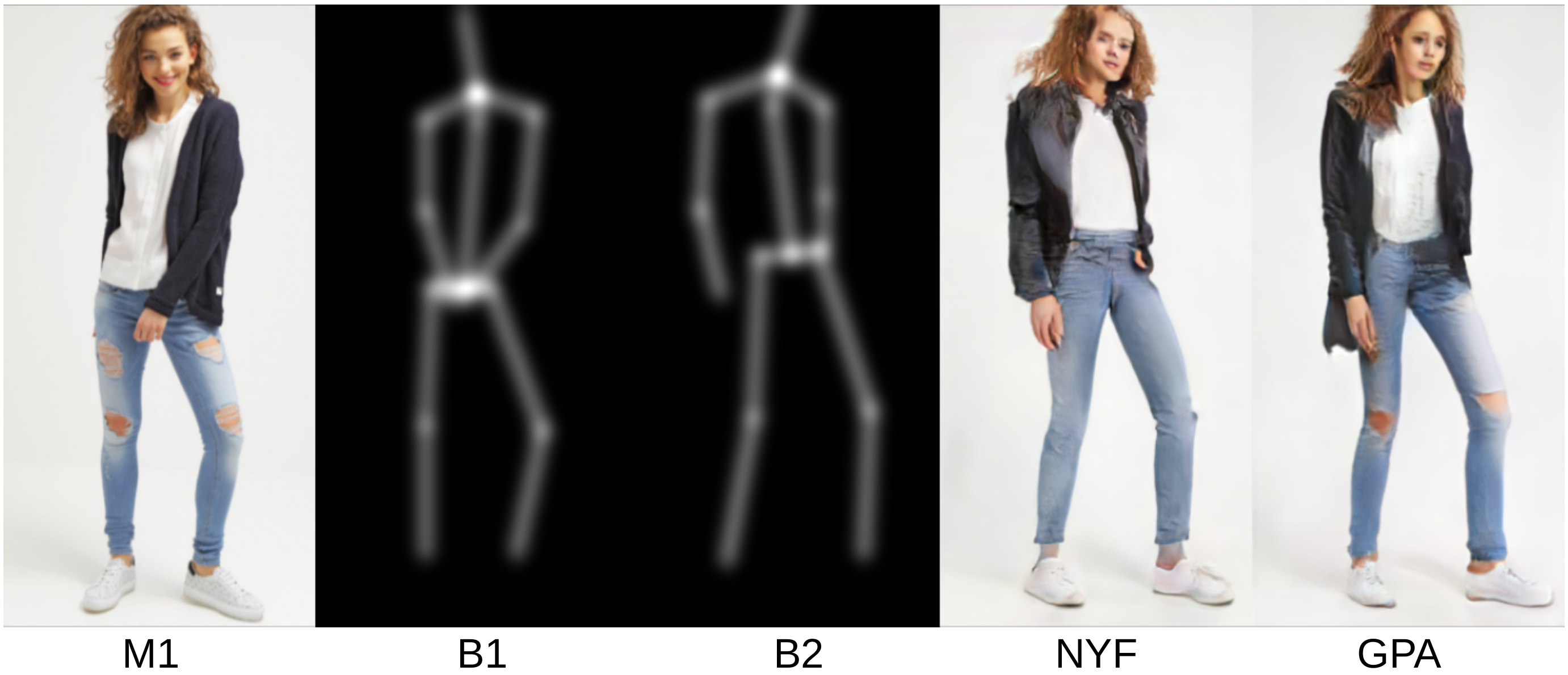}
        \caption{Visual comparison of pose morphing via attention copying with GPA or the NYF modules in the 3 spatially largest decoder layers -- GPA enables fine detail generation, such as the ripped jeans.}
        \label{fig:a2}
\end{figure}

We ran further experiments on the dataset used in \cite{yildirim2019generating}, which has size $512\ttimes256$ pixels and more visual details than DF and ChaL, so it is a good test case for attention copying architectures. 
This dataset has only single unpaired images, so we sampled random poses $B_R$ and used as training loss $\mathbb{E}_{(M_1,B_1,B_R)} \left[\mathcal{L}_{self}(M_1,B_1)+\mathcal{L}_{GAN}(M_1,B_1,B_R)\right]$.
Architecture was the same as for ChaL.
Fig.\  \ref{fig:a2} shows how well GPA recreates details, and avoids the blur of other attention methods (NYF \cite{xiong2021nystromformer}). 

\begin{figure}[tb]
\begin{minipage}[t]{0.49\textwidth}
\includegraphics[height=3.35cm]{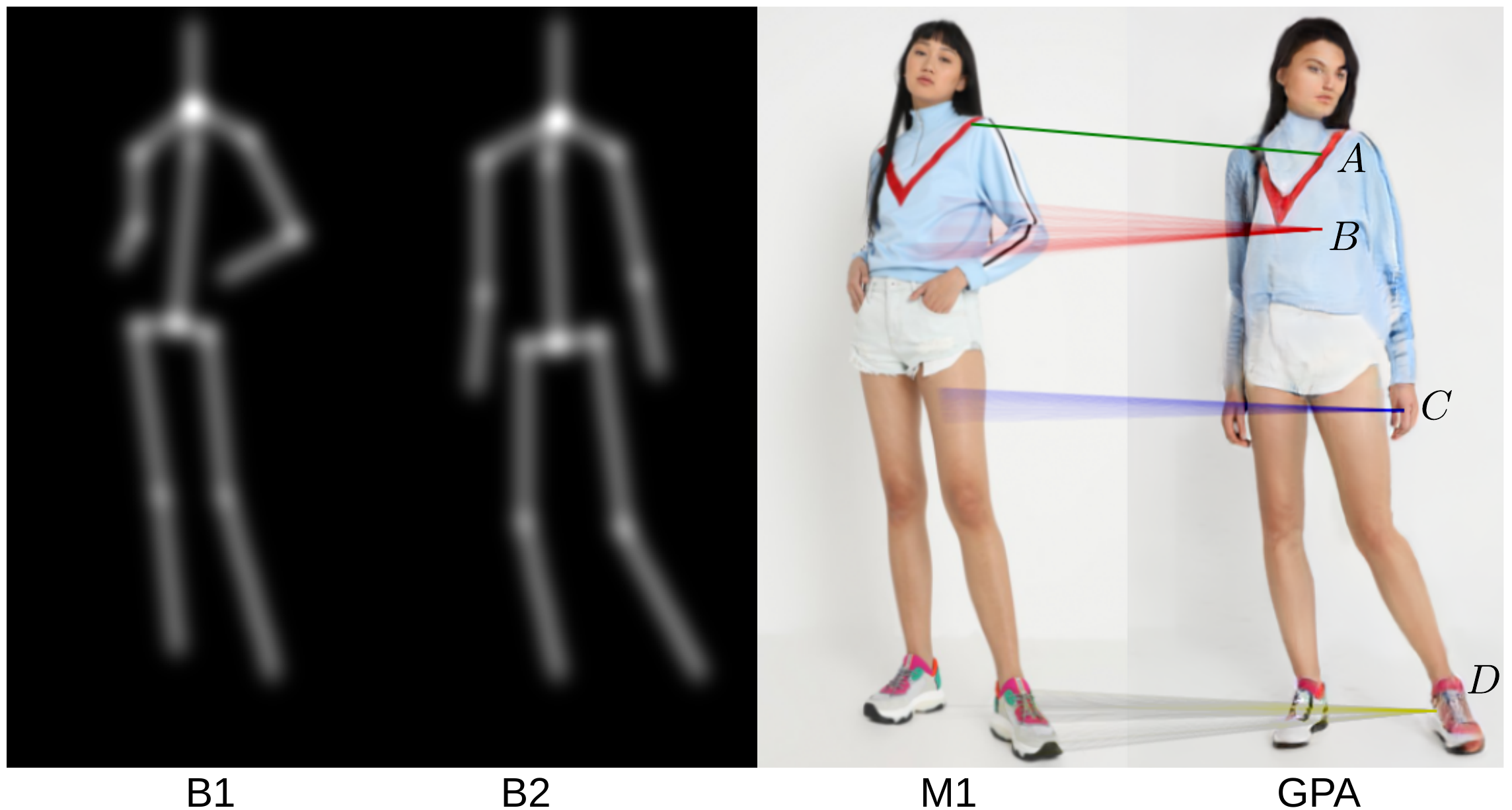}
\captionof{figure}{A visualisation of 4 query positions $A,B,C,D$.
Each is connected with a line to its highest affinity keys.
}
\label{fig:visAtt}
\end{minipage}
\hfill
\begin{minipage}[t]{0.49\textwidth}
\includegraphics[height=3.35cm]{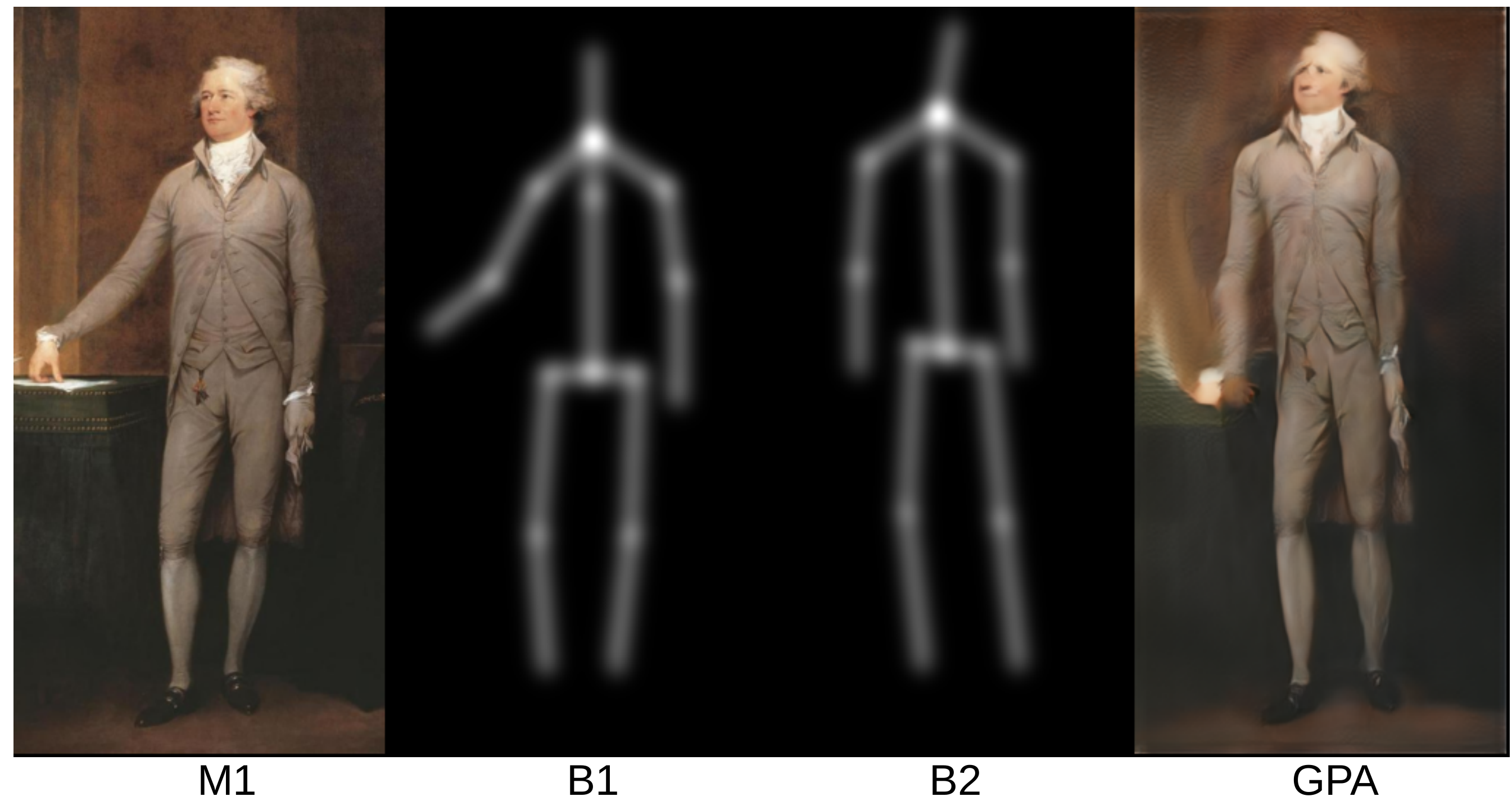}
\captionof{figure}{Example of extreme out-of-sample generation: good transfer capabilities of the architecture using attention copying with GPA.}
\label{fig:hamil}
\end{minipage}
\end{figure}

We examine how GPA picks sparse key sets for attention at a query location, for the largest resolution $512\ttimes256$. 
Fig.\  \ref{fig:visAtt} visualizes the key contributions. 
For a few query locations $q_{i,j}$ we draw lines to keys $k_{i',j'}$, with alpha value proportional to the affinity. We see that for some locations (e.g.\ neck $A$) just a single key contributes the most, i.e. local copying. The shoe $D$ copies information from locations around the shoes in the source appearance image. The blouse $B$ is spread and copies from a larger area with the right color. Interestingly, the hand $C$ is occluded in the source appearance, so GPA learned to fill it with information from the legs, with a similar skin color. 
 Note: attention is performed in feature space, not RGB channels space, but the visualisation uses RGB images as background, in order to show what information the key/query tensors roughly represent.\footnote{Convolutional architectures have local displacement due to the receptive field}

We also tested the limits of the model trained on \cite{yildirim2019generating}, by taking at inference time  a random target pose $B_2$ and a source $M_1,B_1$ taken from a different dataset \cite{MetArt}. Fig.\  \ref{fig:hamil} shows the results -- oil paintings are very different domains than fashion photography, but the copying module can adapt.

\section{Conclusion}
We presented the GPA attention layer: an efficient attention approximation, with hyper-parameters to trade-off memory and accuracy. It is especially useful as copying mechanism in a novel hybrid generative architecture. 
Our experiments showed that attention copying with GPA is a new state-of-the-art method for the pose morphing Deep Fashion benchmark. 
We defined the GPA algorithm for images with 2-d spatial structure, but it is straightforward to adapt similar attention methods to 1-d sequences (e.g. audio data), 3-d (video stream), or even more complicated spatial manifolds (e.g.\ graphs as a generalization of grids) because its basic operators (downsampling, upsampling and partitioning) can be defined on such structures as well.
We expect that GPA is suitable as general attention mechanism for many diverse architectures and tasks, including such with causal masks, but we leave this to future work. 
Currently, Phase 1 of the GPA algorithm is designed to select the top key indices for each query in a non-differentiable manner. 
Recent research on differentiable optimization and sorting \cite{blondel2020fast,cordonnier2021differentiable} makes fully differentiable GPA layers feasible, but this is the topic of future work.


\newpage
\pagebreak 

\bibliographystyle{unsrtnat}
\bibliography{references}


\newpage

\appendix
\begin{appendices}

\section{Additional high resolution results}
\begin{figure}[tb]
\centering
\includegraphics[width=12cm]{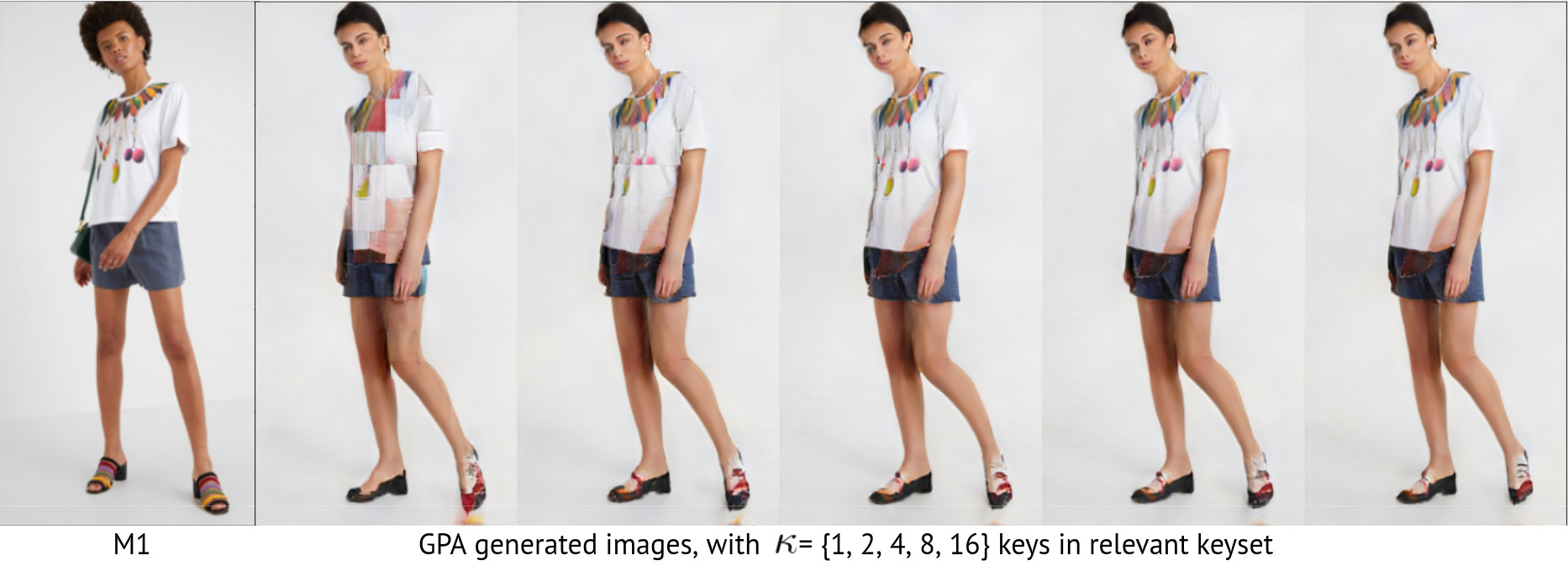}
\caption{Generating the image of the source appearance in a different pose. Increasing the $\kappa$ parameters of the GPA attention trades-off between cheap+approximate attention versus accurate+expensive full attention behaviour. The generated images sorted from left to right with increasing $\kappa$, which correlates with higher visual detail fidelity, e.g.\@ see how the cherry in the logo appears gradually.}
\label{fig:kapabl}
\end{figure}

We include another figure \ref{fig:kapabl} with results for the $\kappa$ ablation (similar to Fig.\ \ref{fig:kapabl2}). We think that the large resolution, larger size (close to 1 millions image) and the nice level of unique details of that fashion dataset \cite{yildirim2019generating} showcases the usefulness of conditioning image copying especially well. It is very difficult to learn from training data only such details as the cherries in the image, but efficient copying from the conditioning image can handle well these unique details seen only at inference time.

\section{Architecture details}

\begin{table}[ht]
  \caption{DF pose morphing architecture -- sizes of Unet decoder blocks.}
  \centering
  \begin{tabular}{lllll}
        \hline
   layer number & spatial size    & channels     & attention type & GPA downsampling $d$  \\
   \hline
   1& $8\ttimes8$ & 640 & full attention & - \\
   2& $16\ttimes16$ & 640 & full attention & -\\
   3& $32\ttimes32$ & 640 & full attention & -\\
   4&$64\ttimes64$ & 640 & full attention & -\\
   5& $128\ttimes128$ & 320 & GPA & 2 \\
   6& $256\ttimes256$ &160 & GPA & 4 \\
    \hline
  \end{tabular}
\label{tab:decoder}
\end{table}

\begin{table}[ht]
  \caption{DF pose morphing architecture -- sizes of Unet encoder blocks, same for Discriminator blocks.}
  \centering
  \begin{tabular}{llll}
        \hline
   layer number & spatial size    & channels     & attention type \\
   \hline
   1& $256\ttimes256$ &160 & - \\
   2& $128\ttimes128$ & 320 & -\\
   3&$64\ttimes64$ & 640 &  -\\
   4& $32\ttimes32$ & 640 & -\\
   5& $16\ttimes16$ & 640 &  -\\
   6& $8\ttimes8$ & 640 & - \\
\hline
  \end{tabular}
\label{tab:encoder}
\end{table}

\begin{figure}[tb]
\centering
\includegraphics[width=6cm]{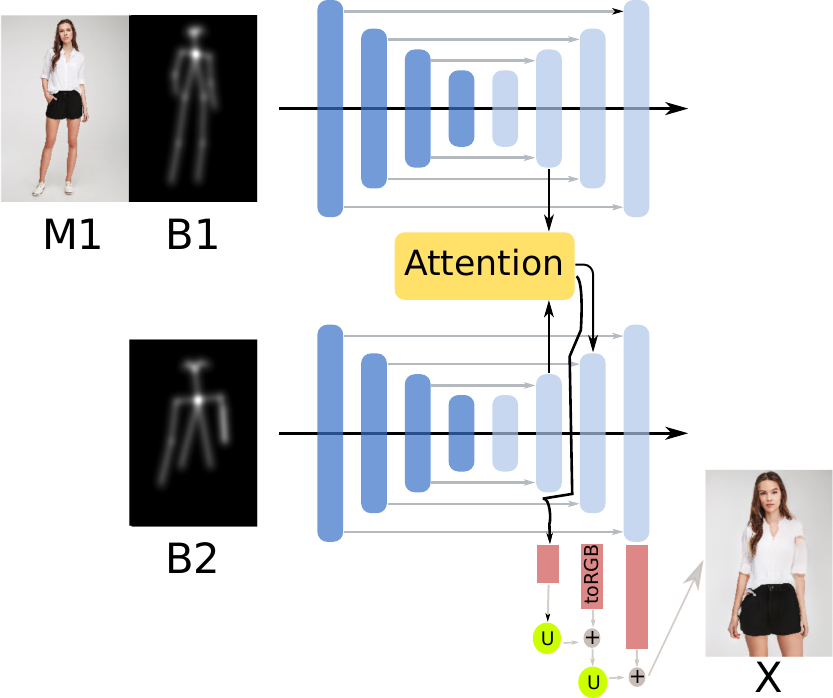}
\caption{ Scheme of pose-morphing generator architecture $G$ with two UNets: one for the concatenated source appearance (M1) and pose (B1) and one for the target pose (B2). The later outputs the generated target appearance (M2), by using a residual-skip scheme to add (upsampled) RGB space outputs, as StyleGAN2 does.}
\label{fig:posearchi3}
\end{figure}

We give in Table \ref{tab:decoder} details for the $256\ttimes256$ pixel architecture for deep fashion generation.
These are the decoder blocks of the unets of the pose morphing with attention copying architecture.
They have also upsampling and LeakyReLU nonlinearities, as in the \cite{karras2020analyzing}.
We also indicate at which levels an approximate attention was used, and on which levels full attention could be used.
In case a GPA approximate attention is used, we give also the downsampling factor, chosen to downsample to $64\ttimes64$ pixels for Phase 1.
As in \cite{karras2020analyzing}, the final RGB space image output is coming from additional convolutions with $c\ttimes3$ channels, projecting feature maps to RGB space, and upsampling and adding them residually for every decoder layer (as in StyleGAN2). This is shown in Fig.\ \ref{fig:posearchi3}.
Decoder blocks use the mean of the last encoder layer as the style input for the StyleGAN blocks.

Encoder blocks are similar, but without attention, and with downsampling instead of upsampling, and without any style input (style vector can be considered constant). Discriminator blocks are identical to encoder blocks, as described in Table  \ref{tab:encoder}.

The architectures for the larger experiments in section \ref{sec:highres} are slightly modified, see Table \ref{tab:decoder2} for the decoder channels and attentions, the encoders and discriminator are similar.

\begin{table}[tb]
  \caption{Pose morphing architecture for 512x256 pixel resolution -- sizes of Unet decoder blocks.}
  \centering
  \begin{tabular}{lllll}
        \hline
   layer number & spatial size    & channels     & attention type & GPA downsampling $d$  \\
   \hline
   1& $16\ttimes8$ & 768 & full attention & - \\
   2& $32\ttimes16$ & 768 & full attention & -\\
   3& $64\ttimes32$ & 768 & full attention & -\\
   4&$128\ttimes64$ & 512 & GPA & 2\\
   5& $256\ttimes128$ & 256 & GPA & 4 \\
   6& $512\ttimes256$ &128 & GPA & 8 \\
    \hline
  \end{tabular}
\label{tab:decoder2}
\end{table}

\section{Crowdsourcing}

\begin{figure}[ht]
\centering
\includegraphics[width=9cm]{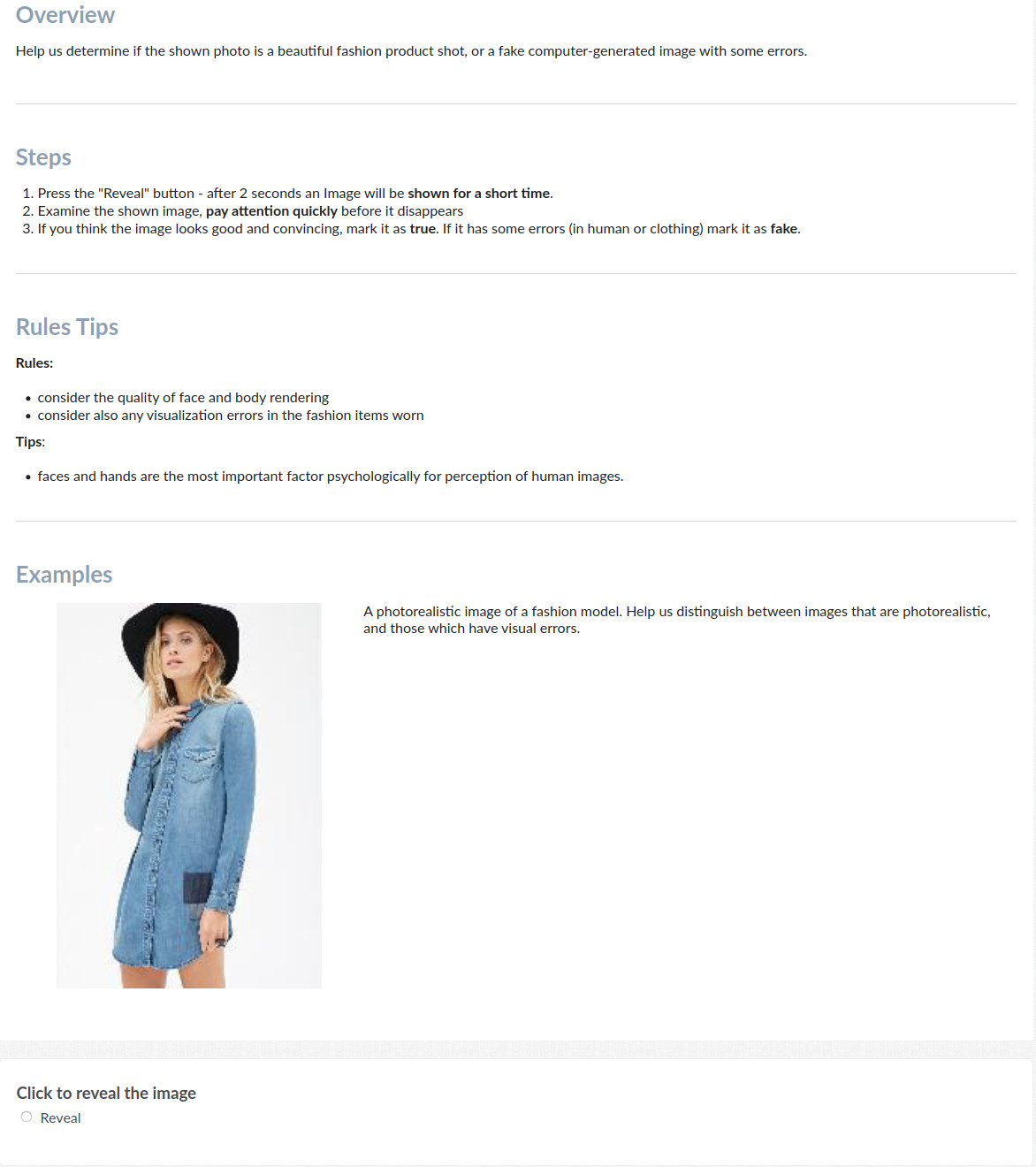}
\caption{Instructions for crowdsourcing perceptual test for DeepFashion pose-morphing.}
\label{fig:crowd_appendix}
\end{figure}

We ran tests using the platform \href{https://client.appen.com}{Figure8appen}, see Fig.\ \ref{fig:crowd_appendix} for example of the setup. An image is shown and the participant has 4 seconds limited decision time to decide "is it true or generated".
We evaluated 4 methods, for each we had 55 fake and 55 true images to evaluate, and gathered 3 clicks per image.
Price was 8 cents per click.
We also had 45 "golden" questions for performance monitoring, requiring participants to score at least 60 percent correctly on these questions.

Given that we expect participants to need (with time switching images) no more than 10 seconds per image, we expect that top workers can get up to 30 dollars per hour, which is a fair wage.
The overall cost for 1320 clicks was around 120 dollars , including overhead for testing and task setting. 

\section{Broader impact}
The GPA quality approximation can allow researchers to train larger attention models than what was possible before, and have smaller computational GPU budgets. This can be a net positive
\begin{itemize}
\item for the limited budgets of small research organisations
\item for the environment by reducing overall electricity consumption.
\end{itemize}

In general, our motivation when designing the experiments was to get the maximal model accuracy on a limited computation budget. E.g. our experiments were always ran on a single V100 GPU, for limited amount of time (5 or 7 days). This is much cheaper computationally than the alternative: running all methods for an unlimited amount of time until convergence, on multiple GPU cores. We think that such an approach keeps deep learning research feasible for the resources of small research labs. 

Application-wise, we have not dealt with any sensitive personal information records, and also the images we have created were solely for the purpose of deep learning architecture benchmarking. However, as a general word of caution, pose morphing is closely related to DeepFake technologies. While we consider static image editing not so problematic, the closely related application of morphing of high resolution video signals can have much larger impact on society.
Research in this area should be open and ethical, to prevent abuses and disinformation.
\end{appendices}
\end{document}